\documentclass[10pt,twocolumn,letterpaper]{article}

\usepackage{cvpr}      
\usepackage{multirow}

\usepackage{algorithm}         
\usepackage[noend]{algpseudocode} 
\usepackage{amsmath}
\usepackage{graphicx}
\usepackage{tabularx}

\definecolor{cvprblue}{rgb}{0.21,0.49,0.74}
\usepackage[pagebackref,breaklinks,colorlinks,allcolors=cvprblue]{hyperref}

\def\paperID{15126} 
\def\confName{CVPR}
\def\confYear{2025}

\title{CoTBox-TTT: Grounding Medical VQA with Visual Chain-of-Thought Boxes During Test-time Training}

\author{
Jiahe Qian$^{1}$\thanks{Equal contribution.} \quad
Yuhao Shen$^{2}$\footnotemark[1] \quad
Zhangtianyi Chen$^{2}$ \quad
Juexiao Zhou$^{2}$\thanks{Corresponding author: juexiao.zhou@gmail.com.} \quad
Peisong Wang$^{1}$\\[4pt]
$^{1}$Institute of Automation, Chinese Academy of Sciences\\
$^{2}$School of Data Science, The Chinese University of Hong Kong, Shenzhen
}
\begin{document}
\maketitle
\begin{abstract}
Medical visual question answering could support clinical decision making, yet current systems often fail under domain shift and produce answers that are weakly grounded in image evidence. This reliability gap arises when models attend to spurious regions and when retraining or additional labels are impractical at deployment time. We address this setting with CoTBox-TTT, an evidence-first test-time training approach that adapts a vision–language model at inference while keeping all backbones frozen. The method updates only a small set of continuous soft prompts. It identifies question-relevant regions through a visual chain-of-thought signal and encourages answer consistency across the original image and a localized crop. The procedure is label free, and plug and play with diverse backbones. Experiments on medical VQA show that the approach is practical for real deployments. For instance, adding CoTBox-TTT to LLaVA increases closed-ended accuracy by 12.3\% on pathVQA.
\end{abstract}    
\section{Introduction}
\label{sec:intro}

Medical visual question answering plays a growing role in clinical decision support and in building trust through transparent interactions \cite{li2023llava,singhal2025toward,moor2023med,thawkar2023xraygpt,bazi2023vision}. Real deployments face substantial domain shift across institutions, devices, acquisition protocols, and patient populations \cite{li2018adaptive}. Tasks are open ended and answers are generated in natural language \cite{zhang2023pmc,lau2018dataset,he2020pathvqa,liu2021slake}. Recent vision language models perform well in distribution but their answers often degrade under shift \cite{karmanov2024efficient,zanella2025realistic,zhang2023domainadaptor}. Models may attend to spurious regions and the generated text can drift or hallucinate \cite{wu2024hallucination,li2023evaluating,zhou2023analyzing,leng2024mitigating}. There is a strong need for an approach that adapts a model to each test case without extra labels and without costly retraining \cite{niu2023towards}. Our goal is to improve answer accuracy, stability, and interpretability at the moment of inference.

Existing strategies are limited. Conventional fine tuning requires supervision and multiple training rounds. Popular test time adaptation methods target discriminative classification and do not directly control the two prerequisites of reliable generation in medical VQA \cite{sun2020test,wang2020tent,liu2021ttt++,wang2022continual,niu2022efficient,zhang2022memo,iwasawa2021test,chen2022contrastive,su2022revisiting,juwon2023leveraging}. A model must first attend to the correct visual evidence \cite{chen2024r,luo2024vividmed,nguyen2025localizing} and then produce an answer that remains consistent when the visual context changes slightly. Directly aligning answers without evidence constraints can reinforce self consistent yet incorrect outputs \cite{goyal2017making,agrawal2018don}. These gaps motivate an evidence first design for test-time training that remains label free and low parameter \cite{lester2021power,zhu2024efficient,li2021prefix,liu2021p}.

Our central intuition is that a strong answer should be grounded in explicit visual evidence and should remain consistent across complementary views of that evidence. We operationalize visual chain of thought as a bounding box that marks the region most helpful for answering the question. For each test case the model first predicts a box on the original image and then validates this evidence on a cropped view of the same region. We enforce self consistency of the two boxes. This anchors attention on the question relevant area and suppresses reliance on spurious cues. We then encourage answer consistency between the original view and the cropped view. The student distribution is guided by an exponential moving average teacher through sequence-level text alignment. The design is effective because it converts correct evidence selection into a direct supervisory signal on generation, which reduces hallucination and improves stability. The entire procedure uses only a small set of continuous soft prompts that are updated at test time.

Our approach is model agnostic and plug and play. The test time training head attaches to diverse vision language backbones without changing encoders or projection layers. The parameter and compute overhead are small which supports near line or online adaptation in clinical workflows. Across multiple medical VQA benchmarks and cross domain settings the method consistently improves accuracy and stability while producing an interpretable evidence trail that links each answer to an explicit region. In summary our work makes three contributions.

\begin{itemize}
\item An evidence driven test time training mechanism that enforces self consistency of visual chain of thought bounding boxes.
\item A cross view answer consistency objective for generative vision language models guided by an exponential moving average teacher.
\item A parameter efficient head that integrates with a wide range of models and requires no reinforcement learning.
\end{itemize}

\section{Related Work}
\label{sec:related}

\subsection{Medical VQA}
Medical visual question answering has progressed from small single-domain settings toward open-ended and clinically oriented scenarios. PMC-VQA establishes large-scale visual instruction data for medical VQA and highlights brittleness in free-form generation under distribution shift \cite{zhang2023pmc}. Biomedical vision–language models such as LLaVA-Med show that instruction tuning can induce strong priors for clinical images, yet hallucination and evidence drift remain open challenges in deployment \cite{li2023llava,wu2024hallucination}. Region-grounded pipelines bring explicit localization into the loop. R-LLaVA injects region-of-interest supervision to better connect textual rationales with image evidence \cite{chen2024r}. VividMed broadens visual grounding operators for fine-grained medical dialogue and question answering \cite{luo2024vividmed}. Grounded evaluation protocols that require localizing before answering help disentangle evidence selection from language generation and diagnose evidence–answer mismatch \cite{nguyen2025localizing}. Clinician-facing studies further suggest that carefully grounded and calibrated systems can assist expert-level medical question answering \cite{singhal2025toward}.

\subsection{Test-Time Training}
Test-time adaptation and training aim to mitigate distribution shift without labeled target data. \cite{sun2020test} introduced a self-supervised auxiliary objective that is optimized on each test sample to improve robustness under shift. Entropy minimization then framed fully test-time adaptation as confidence-driven online optimization and demonstrated substantial gains on corrupted and shifted benchmarks \cite{wang2020tent}. Subsequent analyses clarified failure modes and proposed feature alignment strategies that stabilize adaptation across diverse shifts and batch regimes \cite{liu2021ttt++}. Recent studies have begun to explore test-time procedures for vision–language models with efficiency in mind, including dynamic adapters and prompt-tuning variants that preserve zero-shot competence while reducing computation \cite{karmanov2024efficient,zhu2024efficient}. Together these developments provide a foundation for test-time procedures in multimodal systems and motivate designs that respect the requirements of generation under clinical distribution shift.

\section{Method}

\noindent This section is organized as follows. We first present an overview of CoTBox-TTT that defines the test-time training setting in Section \ref{method:overview}. We then detail self consistency by evidence localization with the grounding model \(\mathcal{G}\) and the validated box loss in Section \ref{method:evidence}. Next we describe cross-view answer consistency with the EMA teacher for \(\mathcal{F}\) and the language modeling objective across views in Section \ref{method:consistency}. We formulate the overall objective and the two step optimization schedule in Section \ref{method:objective}. Finally we provide implementation details including model choices, prompt sizes, initialization, learning rates, and the update schedule in Section \ref{method:implementation}. 

\begin{figure*}[t]  
    \centering
    \includegraphics[width=1\linewidth]{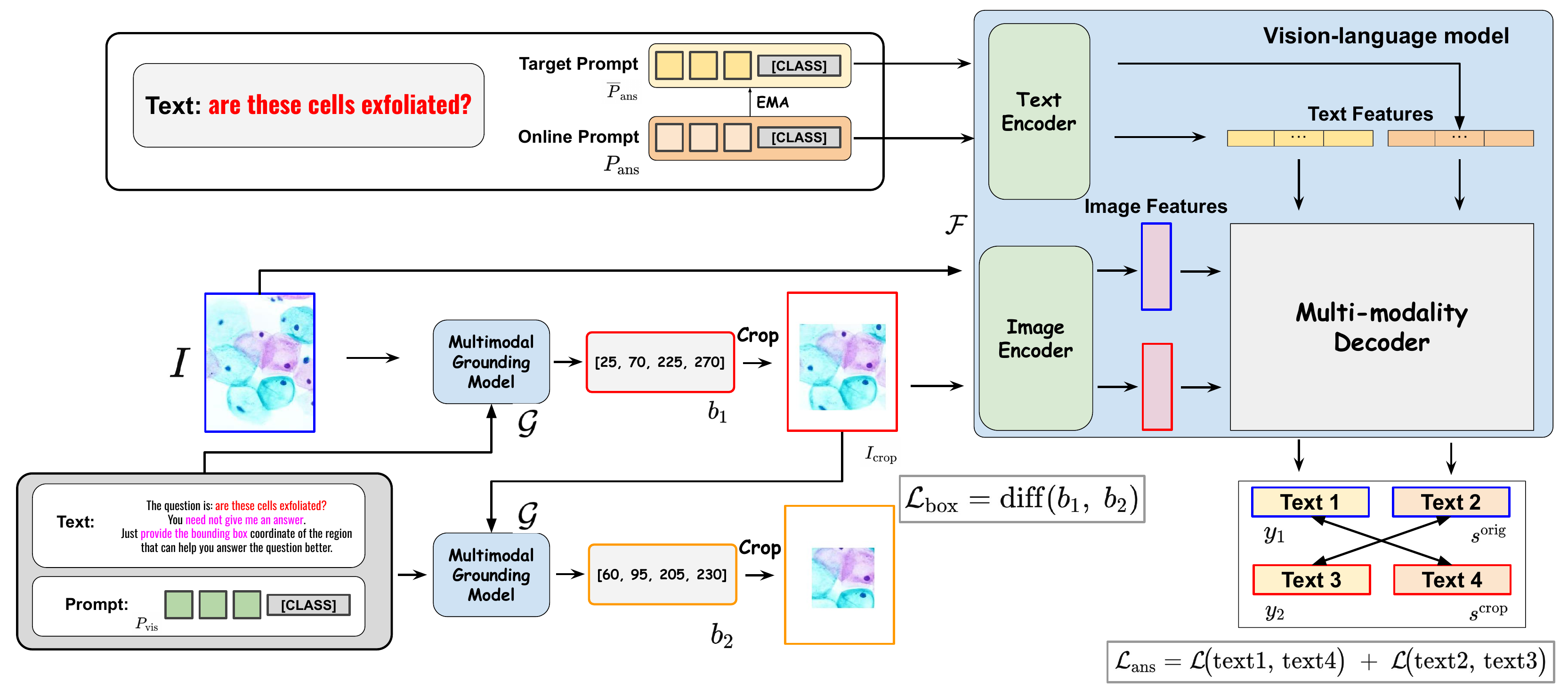}
    \caption{Overview of CoTBox-TTT. (a) Evidence localization: a grounding model conditioned on visual prompts predicts a bounding box on the original image, crops a localized view, and performs a second pass to validate the evidence. (b) Answer consistency: a vision–language model generates student answers on the original and cropped views while an EMA teacher provides targets on the same two views, and the student is aligned to the teacher across views. (c) Test-time adaptation updates only small soft prompts and keeps encoders and decoders frozen, yielding an interpretable evidence trail and consistent gains across backbones.}
    \label{fig:4}
\end{figure*}

\subsection{Overview} \label{method:overview}
CoTBox-TTT operates in a test-time training setting. Given an image-question pair \((I,q)\), all backbone, encoder, and projection parameters remain frozen. Adaptation is restricted to two small sets of continuous soft prompts. The first set \(P_{\mathrm{vis}}\) conditions a pre-trained multi-modality grounding model \(\mathcal{G}\) that emits question-relevant bounding boxes under a fixed JSON schema. The second set \(P_{\mathrm{ans}}\) conditions a vision-language model \(\mathcal{F}\) that generates textual answers. An exponential moving average copy \(\overline{P}_{\mathrm{ans}}\) is maintained as a teacher and does not receive gradients.

As shown in Figure \ref{fig:4}, the procedure contains two stages that share the same frozen backbones. The evidence stage runs \(\mathcal{G}\) twice with shared prompts \(P_{\mathrm{vis}}\). It first predicts a box on the original image and then validates the evidence on the cropped view. A single language modeling loss on the validated box string updates \(P_{\mathrm{vis}}\). The answer stage runs \(\mathcal{F}\) on the original image and on the crop. A single language modeling loss aligns student outputs to teacher sequences across the two views and updates \(P_{\mathrm{ans}}\). The teacher prompts \(\overline{P}_{\mathrm{ans}}\) are refreshed by an exponential moving average. Bounding boxes follow a strict schema and the crop is a deterministic function of the predicted coordinates. The design is fully label free.

\subsection{Self-consistency by Evidence Localization} \label{method:evidence}
Given \((I,q)\), the grounding model \(\mathcal{G}\) is conditioned by the soft prompts \(P_{\mathrm{vis}}\). The two passes share the same \(P_{\mathrm{vis}}\) and all other parameters are frozen. The model outputs a bounding box with absolute pixel coordinates under a fixed JSON schema. The schema is written as \(\{\text{``bbox''}:[x_1,y_1,x_2,y_2]\}\). Let \(T_b\) denote the fixed sequence length under the shared template.

The first pass predicts on the original image
\begin{equation}
b_1 = \mathcal{G}(I, q; P_{\mathrm{vis}}),
\end{equation}
where \(b_1=[x_1,y_1,x_2,y_2]\) are integer pixel coordinates within image bounds. The crop for the second pass is computed by a deterministic operator
\begin{equation}
I_{\mathrm{crop}} = \mathrm{Crop}(I, b_1),
\end{equation}
where \(\mathrm{Crop}(\cdot)\) clips coordinates to the valid pixel range of \(I\), extracts the rectangular region, pads the result back to the original image size using white pixels, and resizes the padded image to the input resolution of \(\mathcal{G}\). The second pass validates the evidence on the crop
\begin{equation}
b_2 = \mathcal{G}(I_{\mathrm{crop}}, q; P_{\mathrm{vis}}),
\end{equation}
where \(b_2=[x'_1,y'_1,x'_2,y'_2]\) follows the same schema. Let \(P_{\mathcal{G}}(\cdot \mid \cdot)\) be the next token conditional probability defined by \(\mathcal{G}\) under \(P_{\mathrm{vis}}\) with teacher forcing on the prefix of the string \(b_2\). CoTBox-TTT uses a single language modeling loss
\begin{equation}
\mathcal{L}_{\mathrm{box}} = - \frac{1}{T_b} \sum_{t=1}^{T_b} \log P_{\mathcal{G}}\!\left(b_{2,t} \mid b_{2,<t}, I, q; P_{\mathrm{vis}}\right),
\end{equation}
where \(b_{2,t}\) is the target token at position \(t\) in the string \(b_2\) and \(b_{2,<t}\) is its prefix. 

\subsection{Cross-view Consistency with EMA Teacher} \label{method:consistency}
Given \((I,q)\) and the crop \(I_{\mathrm{crop}}\), the answer model \(\mathcal{F}\) is conditioned by the soft prompts \(P_{\mathrm{ans}}\). The network weights are shared across branches and remain frozen. Let \(\overline{P}_{\mathrm{ans}}\) be an exponential moving average copy of \(P_{\mathrm{ans}}\) that serves as a teacher. At the beginning of an adaptation episode the teacher is initialized to the current student prompts.

Student outputs are generated on both views
\begin{equation}
\begin{aligned}
y_1 &= \mathcal{F}(I, q; P_{\mathrm{ans}}),\\
y_2 &= \mathcal{F}(I_{\mathrm{crop}}, q; P_{\mathrm{ans}}),
\end{aligned}
\end{equation}
where \(y_1\) and \(y_2\) are token sequences produced by teacher forced decoding on the corresponding inputs. Teacher sequences are produced with the EMA prompts
\begin{equation}
\begin{aligned}
s^{\mathrm{orig}} &= \mathcal{F}(I, q; \overline{P}_{\mathrm{ans}}),\\
s^{\mathrm{crop}} &= \mathcal{F}(I_{\mathrm{crop}}, q; \overline{P}_{\mathrm{ans}}),
\end{aligned}
\end{equation}
where the lengths are \(T_{\mathrm{orig}}\) and \(T_{\mathrm{crop}}\). Let \(P_{\mathcal{F}}(\cdot \mid \cdot)\) denote the next–token conditional probability defined by \(\mathcal{F}\) under the given prompts. We compute the cross–view objective as the mean of two view–specific language modeling losses. The first loss supervises the student with the teacher sequence on the original view
\begin{equation}
\mathcal{L}_{\mathrm{ans}}^{\mathrm{orig}}
=
- \frac{1}{T_{\mathrm{orig}}}
\sum_{t=1}^{T_{\mathrm{orig}}}
\log P_{\mathcal{F}}\!\left(
s^{\mathrm{orig}}_{t}\mid s^{\mathrm{orig}}_{<t},\, I,\, q;\, P_{\mathrm{ans}}
\right),
\end{equation}
where \(s^{\mathrm{orig}}_{t}\) is the teacher token at position \(t\) and \(s^{\mathrm{orig}}_{<t}\) is its prefix. The second loss supervises the student with the teacher sequence on the cropped view
\begin{equation}
\mathcal{L}_{\mathrm{ans}}^{\mathrm{crop}}
=
- \frac{1}{T_{\mathrm{crop}}}
\sum_{t=1}^{T_{\mathrm{crop}}}
\log P_{\mathcal{F}}\!\left(
s^{\mathrm{crop}}_{t}\mid s^{\mathrm{crop}}_{<t},\, I_{\mathrm{crop}},\, q;\, P_{\mathrm{ans}}
\right),
\end{equation}
where \(s^{\mathrm{crop}}_{t}\) is the teacher token at position \(t\) and \(s^{\mathrm{crop}}_{<t}\) is its prefix. 

The overall cross–view objective is the sum of the two view–specific losses
\begin{equation}
\mathcal{L}_{\mathrm{ans}}
=
\mathcal{L}_{\mathrm{ans}}^{\mathrm{orig}}
+
\mathcal{L}_{\mathrm{ans}}^{\mathrm{crop}}.
\end{equation}


\subsection{Overall Objective and Algorithmic Scheme} \label{method:objective}
This subsection specifies the joint objective used by CoTBox-TTT and how it is applied during test-time adaptation. The procedure is shown in Algorithm \ref{alg:cotbox_concise}

The total objective is written as
\begin{equation}
\mathcal{L}_{\mathrm{total}}=\alpha\,\mathcal{L}_{\mathrm{box}}+\beta\,\mathcal{L}_{\mathrm{ans}},
\end{equation}
where \(\alpha>0\) and \(\beta>0\) are scalar weights. In practice CoTBox-TTT applies a two-step schedule that decouples the updates.

During the evidence step the prompts \(P_{\mathrm{vis}}\) are updated by minimizing the evidence loss while all other parameters are frozen
\begin{equation}
\min_{P_{\mathrm{vis}}}\ \mathcal{L}_{\mathrm{box}}(I,q;\mathcal{G},P_{\mathrm{vis}}).
\end{equation}
After the update the crop \(I_{\mathrm{crop}}=\mathrm{Crop}(I,b_1)\) is fixed for the next step
\begin{equation}
(\alpha,\beta)=(1,0).
\end{equation}
The gradient of \(\mathcal{L}_{\mathrm{box}}\) with respect to \(P_{\mathrm{ans}}\) is zero by construction
\begin{equation}
\frac{\partial \mathcal{L}_{\mathrm{box}}}{\partial P_{\mathrm{ans}}}=0.
\end{equation}

During the answer step the prompts \(P_{\mathrm{ans}}\) are updated by minimizing the cross-view loss while all other parameters are frozen
\begin{equation}
\min_{P_{\mathrm{ans}}}\ \mathcal{L}_{\mathrm{ans}}(I,I_{\mathrm{crop}},q;\mathcal{F},P_{\mathrm{ans}},\overline{P}_{\mathrm{ans}}).
\end{equation}
The schedule sets
\begin{equation}
(\alpha,\beta)=(0,1).
\end{equation}
No gradient flows from the answer loss to the evidence prompts
\begin{equation}
\frac{\partial \mathcal{L}_{\mathrm{ans}}}{\partial P_{\mathrm{vis}}}=0.
\end{equation}

For completeness the per-step updates follow standard first-order optimization on the prompts. The evidence update is
\begin{equation}
P_{\mathrm{vis}} \leftarrow P_{\mathrm{vis}} - \eta_{\mathrm{vis}} \,\nabla_{P_{\mathrm{vis}}}\mathcal{L}_{\mathrm{box}},
\end{equation}
where \(\eta_{\mathrm{vis}}>0\) is the learning rate. The answer update is
\begin{equation}
P_{\mathrm{ans}} \leftarrow P_{\mathrm{ans}} - \eta_{\mathrm{ans}} \,\nabla_{P_{\mathrm{ans}}}\mathcal{L}_{\mathrm{ans}},
\end{equation}
where \(\eta_{\mathrm{ans}}>0\) is the learning rate. The teacher prompts are refreshed by an exponential moving average
\begin{equation}
\overline{P}_{\mathrm{ans}} \leftarrow \beta\,\overline{P}_{\mathrm{ans}} + \left(1-\beta\right) P_{\mathrm{ans}},
\end{equation}
where \(\beta \in (0,1)\) is the decay factor. This schedule yields a total objective that is simple to state yet applied in two sequential minimizations with disjoint gradient flow.

\begin{algorithm}[t]
\caption{CoTBox-TTT procedure}
\label{alg:cotbox_concise}
\begin{algorithmic}[1]
\Require Image–question $(I,q)$, grounding model $\mathcal{G}$, answer model $\mathcal{F}$, prompts $P_{\mathrm{vis}}, P_{\mathrm{ans}}$, EMA decay $\beta$, mini epochs $E$
\State $\overline{P}_{\mathrm{ans}} \gets P_{\mathrm{ans}}$
\For{$e=1$ to $E$}
  \State $b_1 \gets \mathcal{G}(I,q;P_{\mathrm{vis}})$
  \State $I_{\mathrm{crop}} \gets \mathrm{Crop}(I,b_1)$
  \State $b_2 \gets \mathcal{G}(I_{\mathrm{crop}},q;P_{\mathrm{vis}})$
  \State $P_{\mathrm{vis}} \gets \mathrm{SGD}\!\left(P_{\mathrm{vis}}, \nabla \mathcal{L}_{\mathrm{box}}(b_2 \mid I,q)\right)$
  \State $s^{\mathrm{orig}} \gets \mathcal{F}(I,q;\overline{P}_{\mathrm{ans}})$
  \State $s^{\mathrm{crop}} \gets \mathcal{F}(I_{\mathrm{crop}},q;\overline{P}_{\mathrm{ans}})$
  \State $P_{\mathrm{ans}} \gets \mathrm{SGD}\!\left(P_{\mathrm{ans}}, \nabla \mathcal{L}_{\mathrm{ans}}(s^{\mathrm{orig}},s^{\mathrm{crop}} \mid I,I_{\mathrm{crop}},q)\right)$
  \State $\overline{P}_{\mathrm{ans}} \gets \beta\,\overline{P}_{\mathrm{ans}} + (1-\beta)\,P_{\mathrm{ans}}$
\EndFor
\State \Return final answer $\mathcal{F}(I,q;P_{\mathrm{ans}})$ and box $b_1$
\end{algorithmic}
\end{algorithm}

\subsection{Implementation Details} \label{method:implementation}
The grounding model is VisCoT \cite{shao2024visual} with weights frozen. The answer model, such as LLaVA-Med, keeps its visual encoder, projection layers, and language backbone frozen. Adaptation uses two continuous soft prompts inserted as prefix embeddings and initialized to zero, with 24 tokens for evidence and 32 tokens for answers. The tokenizer and vocabulary are shared across original and cropped views. The maximum answer length is 128 tokens and the tokenized evidence string uses a fixed padded length of 32 tokens for stable loss scaling. Teacher decoding is deterministic with unit temperature and greedy selection. Learning rates are 1e-3 for evidence prompts and 5e-4 for answer prompts. The exponential moving average teacher uses a decay of 0.9 and is synchronized to the student at the start of each adaptation episode. Training follows a two step schedule with 20 epochs per image, where each mini epoch first updates evidence prompts with the evidence loss and then updates answer prompts with the cross view loss. Computation uses eight RTX 4090 GPUs. Other hyperparameters follow backbone defaults and remain unchanged during test time training.

\section{Experiments}

\subsection{Settings}

\noindent\textbf{Baselines}
\begin{itemize}
\item \textit{LLaVA \cite{liu2023visual}}. A general-purpose vision-language assistant trained by visual instruction tuning that connects a CLIP-based image encoder with a large language model. The baseline follows the official inference recipe including image resolution, decoding temperature, maximum answer length, and tokenization. 
\item \textit{LLaVA-Med \cite{li2023llava}}. A biomedical adaptation of LLaVA trained with a two-stage curriculum. Stage one aligns biomedical concepts using large-scale PubMed Central figure–caption pairs. Stage two performs instruction tuning on GPT-4 generated multimodal conversations. The public release includes checkpoints and evaluation scripts for three medical VQA benchmarks. 
\item \textit{Hulu-Med \cite{jiang2025hulu}}. A transparent generalist medical vision-language model that unifies text, 2D images, 3D volumes, and videos through a patch-based vision encoder and an LLM decoder. The official resources provide inference pipelines across 30 medical benchmarks. For comparability here only the 2D VQA setting is considered. 
\end{itemize}

\noindent\textbf{Datasets and Tasks}
\begin{itemize}
\item \textit{VQA-RAD \cite{lau2018dataset}}. A clinician-authored radiology VQA benchmark with 315 images and 3{,}515 question–answer pairs. Questions cover close-ended types such as yes–no and open-ended free-form answers. The standard split and task protocol are used to ensure consistency with prior medical VQA practice. 
\item \textit{SLAKE \cite{liu2021slake}}. A bilingual medical VQA dataset with physician-verified annotations, comprehensive semantic labels, and an associated medical knowledge base. Images span multiple radiology modalities and body regions. The English subset is typically adopted for head-to-head comparisons and both open-ended and close-ended settings are reported. 
\item \textit{PathVQA \cite{he2020pathvqa}}. A pathology VQA dataset with 4{,}998 images and 32{,}799 questions collected from textbooks and a public digital library. The dataset mixes open-ended questions and close-ended yes–no items. The commonly used evaluation split is followed and questions are grouped under the open-ended and close-ended protocols. 
\end{itemize}

\noindent\textbf{Evaluation Metrics}
\begin{itemize}
\item \textit{Close-ended}. Accuracy is computed by exact match against the reference answer set for yes–no and other close-ended types. 
\item \textit{Open-ended}. Recall is computed for free-form generation by checking whether gold keywords appear in the generated answer. Main tables are reported in recall to match established protocols. F1 is additionally provided in the appendix when applicable. 
\end{itemize}

\noindent For each baseline model results are reported under two conditions on all datasets and task types. The first condition is the native inference without CoTBox-TTT. The second condition augments inference with CoTBox-TTT while keeping preprocessing and decoding identical. This design isolates the contribution of CoTBox-TTT under the open-ended and close-ended metrics on VQA-RAD, SLAKE and PathVQA. The
data statistics are provided in Table \ref{tab:dataset_stats}.

\begin{table}[t]
\centering
\begingroup
\setlength{\tabcolsep}{3pt}
\fontsize{8}{10}\selectfont
\begin{tabular}{lccccccccc}
\toprule
\multirow{2}{*}{Dataset} & \multicolumn{2}{c}{VQA-RAD} & \multicolumn{3}{c}{SLAKE} & \multicolumn{3}{c}{PathVQA} \\
\cmidrule(lr){2-3}\cmidrule(lr){4-6}\cmidrule(lr){7-9}
 & Train & Test & Train & Val & Test & Train & Val & Test \\
\midrule
\# Images   & 313 & 203 & 450 & 96  & 96  & 2{,}599  & 858  & 858  \\
\# QA Pairs & 1{,}797 & 451 & 4{,}919 & 1{,}053 & 1{,}061 & 19{,}755 & 6{,}279 & 6{,}761 \\
\# Open     & 770 & 179 & 2{,}976 & 631 & 645 & 9{,}949 & 3{,}144 & 3{,}370 \\
\# Closed   & 1{,}027 & 272 & 1{,}943 & 422 & 416 & 9{,}806 & 3{,}135 & 3{,}391 \\
\bottomrule
\end{tabular}
\caption{Dataset statistics for medical VQA datasets.}
\label{tab:dataset_stats}
\endgroup
\end{table}

\subsection{Main Results}
We compare each backbone in two conditions that differ only by the presence of CoTBox\textendash TTT. Decoding temperature, maximum answer length, preprocessing, and scoring are matched across conditions. Performance is reported on VQA\textendash RAD, SLAKE, and PathVQA under open\textendash ended recall and close\textendash ended accuracy.

CoTBox\textendash TTT improves every metric entry in Table~\ref{tab:main_results}. All forty eight backbone\textendash dataset\textendash metric cells show positive shifts. This holds for generic instruction models and for medically adapted models and for generalist medical models. The gains are largest where open\textendash domain generation is weak at baseline and remain consistently positive on strong backbones. For example LLaVA shows a mean open recall improvement of 11.05 points and a mean closed accuracy improvement of 9.21 points across the three datasets. LLaVA\textendash Med with a CLIP vision encoder improves open recall by 4.75 on VQA\textendash RAD, 4.41 on SLAKE, and 2.68 on PathVQA, with closed accuracy gains of 2.94, 4.32, and 1.50. Hulu\textendash Med\textendash 14B improves open recall by 4.73 on VQA\textendash RAD, 3.31 on SLAKE, and 4.71 on PathVQA, and improves closed accuracy by 5.89, 3.12, and 2.59.

Closed accuracy increases are consistent on all datasets and backbones, which indicates that cross\textendash view alignment stabilizes classification\textendash style questions. On VQA\textendash RAD LLaVA closed accuracy rises from 65.07 to 73.16 and Hulu\textendash Med\textendash 14B rises from 82.35 to 88.24. On SLAKE LLaVA closed accuracy rises from 63.22 to 70.43 and LLaVA\textendash Med\textendash Vicuna rises from 83.17 to 87.26. On PathVQA LLaVA closed accuracy rises from 63.20 to 75.52 and LLaVA\textendash Med\textendash BioMedCLIP rises from 91.09 to 93.02.

Open\textendash ended recall also improves across the board, which shows that the evidence\textendash driven procedure strengthens free\textendash form generation. The largest shift appears on PathVQA for LLaVA where open recall rises from 7.74 to 32.60. Strong models benefit as well. Hulu\textendash Med\textendash 32B on SLAKE rises from 85.06 to 90.26 and LLaVA\textendash Med\textendash BioMedCLIP on SLAKE rises from 87.11 to 90.24. On VQA\textendash RAD the open recall gains are 5.44 for LLaVA and 3.41 to 4.75 across LLaVA\textendash Med variants and 2.26 to 5.73 across Hulu\textendash Med variants.

Taken together these results indicate that CoTBox\textendash TTT acts as a model\textendash agnostic adapter that consistently improves answer reliability. It improves closed accuracy by stabilizing cross\textendash view predictions and improves open recall by grounding free\textendash form generation on explicit visual evidence.

\begin{figure*}[t]  
    \centering
    \includegraphics[width=1\linewidth]{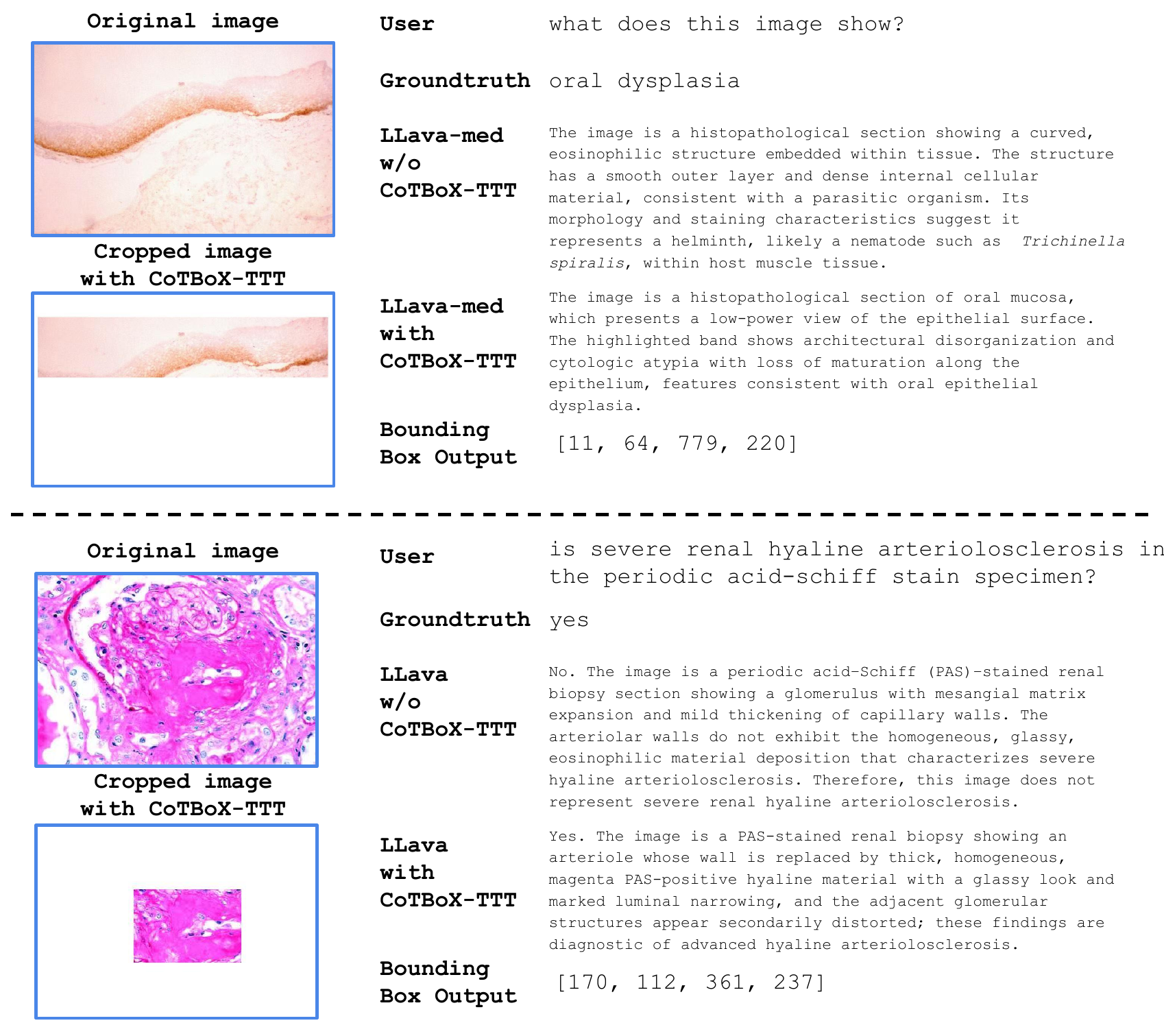}
    \caption{Qualitative examples of CoTBox-TTT. Each case shows the original image, the crop guided by the grounding model with the predicted bounding box, the model responses from baseline model with and without CoTBox-TTT, and the ground truth label. In each case the baseline model without test-time adaptation produces an answer that is either incomplete or inconsistent with the image content, while CoTBox-TTT first localizes the clinically relevant region and then aligns the answers across the original and cropped views in order to convert ambiguous or incorrect predictions into medically specific and image supported statements.}
    \label{fig:5}
\end{figure*}

Figure \ref{fig:5} shows some examples. In the first example the baseline answers are vague and drift toward nonspecific mucosal changes, whereas CoTBox-TTT guides the model to the dysplastic epithelial strip along the mucosal edge and produces the correct diagnosis of oral epithelial dysplasia. The bounding box constrains attention to the atypical epithelium with nuclear crowding and loss of maturation and the crop suppresses distracting keratin debris in the surrounding field. In the second example the PAS stained renal biopsy contains a thickened arteriole with a glassy eosinophilic wall. The baseline mixes interstitial findings with vascular changes and fails to name the vascular lesion, while CoTBox-TTT places the box on the arteriole, attends to the concentric hyaline deposition, and outputs severe hyaline arteriolosclerosis. Across both cases the evidence-guided crop and the cross-view alignment reduce spurious focus and convert ambiguous descriptions into specific clinicopathologic terms that match the ground truth.

\begin{table*}[t]
\centering
\begin{tabular}{l|c|cc|cc|cc}
\toprule
\multirow{2.5}{*}{Model} & \multirow{2.5}{*}{CoTBox\textendash TTT} 
& \multicolumn{2}{|c|}{VQA\textendash RAD} & \multicolumn{2}{|c|}{SLAKE} & \multicolumn{2}{|c}{PathVQA} \\
\cmidrule(lr){3-4}\cmidrule(lr){5-6}\cmidrule(lr){7-8}
& & Open $\uparrow$ & Closed $\uparrow$ & Open $\uparrow$ & Closed $\uparrow$ & Open $\uparrow$ & Closed $\uparrow$ \\
\midrule
\multirow{2}{*}{LLaVA}   &             &   50.00         &    65.07    &   78.18     &   63.22   &   7.74     &   63.20     \\
                         & $\checkmark$&    \textbf{55.44}    &   \textbf{73.16}     &   \textbf{81.04}     &    \textbf{70.43}    &    \textbf{32.60}    &    \textbf{75.52}    \\
\midrule
\multirow{2}{*}{LLaVA\textendash Med\textendash CLIP}            &             &   61.52 & 84.19 & 83.08 & 85.34 & 37.95 & 91.21 \\
                         & $\checkmark$&    \textbf{66.27}    &    \textbf{87.13}    &    \textbf{87.49}    &    \textbf{89.66}    &  \textbf{40.63}      &    \textbf{92.71}    \\
\midrule
\multirow{2}{*}{LLaVA\textendash Med\textendash Vicuna}           &             &   64.39 & 81.98 & 84.71 & 83.17 & 38.87 & 91.65 \\
                         & $\checkmark$&    \textbf{68.25}    &    \textbf{86.40}    &   \textbf{88.12}     &    \textbf{87.26}    &    \textbf{42.90}    &      \textbf{93.03}  \\
\midrule
\multirow{2}{*}{LLaVA\textendash Med\textendash BioMedCLIP}      &             &   64.75 & 83.09 & 87.11 & 86.78 & 39.60 & 91.09 \\
                         & $\checkmark$&    \textbf{68.16}    &    \textbf{87.13}    &   \textbf{90.24}     &   \textbf{90.39}     &   \textbf{42.73}     &    \textbf{93.02}    \\
\midrule
\multirow{2}{*}{Hulu\textendash Med\textendash 7B}   &             &   63.73 & 87.50 & 84.27 & 90.87 & 38.41 & 92.63 \\
                         & $\checkmark$&   \textbf{69.46}     &   \textbf{90.07}     &   \textbf{88.33}     &   \textbf{92.07}     &   \textbf{41.64}     &    \textbf{93.34}    \\
\midrule
\multirow{2}{*}{Hulu\textendash Med\textendash 14B}  &             &   66.92 & 82.35 & 86.20 & 87.02 & 39.34 & 89.37 \\
                         & $\checkmark$ &    \textbf{71.65}    &        \textbf{88.24}    &    \textbf{89.51}    &   \textbf{90.14}    &   \textbf{44.05}     &    \textbf{91.96}    \\
\midrule
\multirow{2}{*}{Hulu\textendash Med\textendash 32B}  &             &   71.11 & 88.24 & 85.06 & 86.54 & 44.37 & 90.08 \\
                         & $\checkmark$&   \textbf{73.37}     &   \textbf{90.07}     &    \textbf{90.26}    &   \textbf{89.42}     &   \textbf{46.04}     &    \textbf{92.22}    \\
\bottomrule
\end{tabular}
\caption{Main results on VQA\textendash RAD, SLAKE, and PathVQA. Open denotes open\textendash ended Recall. Closed denotes close\textendash ended Accuracy. The CoTBox\textendash TTT column indicates whether the model uses CoTBox\textendash TTT at inference.}
\label{tab:main_results}
\end{table*}

\subsection{Ablation Study}

We study four settings on VQA\textendash RAD for each backbone. Row one enables both components and is the full CoTBox\textendash TTT. Row two enables evidence consistency and disables the EMA teacher where the teacher parameters are tied to the student. Row three enables the EMA teacher and replaces the two pass localization with a single pass grounding so there is no second pass to enforce evidence consistency. Row four disables both components and is the native model without test time training.

Both modules contribute and they do so in complementary ways. On LLaVA the full method improves from 50.00 to 55.44 for open and from 65.07 to 73.16 for closed. The evidence only setting reaches 51.77 and 68.01 which confirms that consistent evidence improves both recall and accuracy. The EMA only setting reaches 53.20 and 70.59 which shows that cross view guidance improves both metrics even when localization uses a single pass. On LLaVA\textendash Med\textendash CLIP the full method reaches 66.27 and 87.13 while evidence only reaches 64.30 and 86.03 and EMA only reaches 64.46 and 85.66. On Hulu\textendash Med\textendash 7B the full method reaches 69.46 and 90.07 while evidence only reaches 65.52 and 88.24 and EMA only reaches 65.85 and 88.97.

Taken together these comparisons indicate that the evidence module strengthens grounding which translates into higher recall and accuracy and that the EMA teacher stabilizes answer alignment which further improves both metrics. The full configuration outperforms either single component on LLaVA by 2.24 open and 2.57 closed over EMA only and on Hulu\textendash Med\textendash 7B by 3.61 open and 1.10 closed over EMA only. Similar margins appear on LLaVA\textendash Med\textendash CLIP. These patterns support the view that evidence consistency and EMA guidance address complementary error modes and that combining them yields additive gains.

\begin{table*}[t]
\centering
\begin{tabular}{l|cc|cc}
\toprule
\multirow{2.5}{*}{Model} & \multicolumn{2}{c|}{Components} & \multicolumn{2}{c}{Subset} \\
\cmidrule(lr){2-3}\cmidrule(lr){4-5}
 & Evidence Consistency & EMA Teacher & Open$\uparrow$ & Closed$\uparrow$ \\
\midrule
\multirow{4}{*}{LLaVA}
  & \textbf{\checkmark} & \textbf{\checkmark} & \textbf{55.44} & \textbf{73.16} \\
  & \checkmark &            & 51.77 & 68.01      \\
  &            & \checkmark & 53.20 & 70.59      \\
  &            &            & 50.00 & 65.07 \\
\midrule
\multirow{4}{*}{LLaVA\textendash Med\textendash CLIP}
  & \textbf{\checkmark} & \textbf{\checkmark} & \textbf{66.27} & \textbf{87.13} \\
  & \checkmark &            & 64.30 & 86.03 \\
  &            & \checkmark & 64.46 & 85.66 \\
  &            &            & 61.52 & 84.19 \\
\midrule
\multirow{4}{*}{Hulu\textendash Med\textendash 7B}
  & \textbf{\checkmark} & \textbf{\checkmark} & \textbf{69.46} & \textbf{90.07} \\
  & \checkmark &            & 65.52 & 88.24 \\
  &            & \checkmark & 65.85 & 88.97 \\
  &            &            & 63.73 & 87.50 \\
\bottomrule
\end{tabular}
\caption{Ablation on VQA\textendash RAD. Each backbone is evaluated with controlled toggles for the two CoTBox\textendash TTT components. The first row per block is the full method (both components enabled), the second removes the validation pass, the third removes the EMA teacher, and the fourth is the native model without test\textendash time training. Metrics are open\textendash ended recall and close\textendash ended accuracy.}
\label{tab:ablation_vqarad}
\end{table*}

\section{Discussion}
CoTBox-TTT is a test-time training framework for medical-VQA that is model-agnostic and plug-and-play. The method freezes all backbones and adapts only small sets of soft prompts. It turns explicit visual evidence and aligns answers across the original image and a localized crop with a teacher based on exponential moving average. Across three datasets and multiple backbones it improves accuracy while providing an interpretable evidence trail that links each answer to a concrete region.

Building on these gains, we see three natural directions that follow the same principles of evidence grounding, cross-view agreement, and parameter-efficient adaptation. The first direction is to replace box consistency with mask consistency. Pixel level masks capture shape and boundary and can represent multi focal or diffuse patterns that a rectangle cannot, as shown by strong medical segmentation baselines \cite{ronneberger2015u,isensee2021nnu}. Consistency can be defined on masks predicted in the original view and in the localized view and mapped by the same crop and padding operator, while promptable segmenters provide a practical mechanism for conditioning the evidence on the question \cite{kirillov2023segment,ma2024segment}. Weakly supervised consistency objectives can be transferred to the mask domain to reduce annotation demands and to stabilize updates with very few adaptation steps \cite{chen2021semi}. These ingredients can be combined within the current two stage schedule so that the update remains focused on soft prompts.

A second direction is to introduce adaptive momentum for the teacher. A fixed decay can be too rigid when the student changes at different speeds across cases. Mean teacher style targets have proven effective for stabilizing semi supervised training and suggest that teacher dynamics matter for generalization \cite{tarvainen2017mean}. Momentum encoders in self supervised learning also highlight the importance of well tuned temporal smoothing and temperature schedules \cite{he2020momentum,caron2021emerging}. Recent analyses of diffusion training further show that revisiting or scheduling EMA can improve stability without increasing annotation cost \cite{karras2024analyzing}. Adapting these insights to test time training suggests scheduling momentum across mini epochs and modulating it by uncertainty or student teacher disagreement.

A third direction is to extend CoTBox-TTT to three dimensional and to other domains. For CT or MRI volumes the evidence and answer can be aligned across neighboring slices with architectures that already support volumetric features \cite{hatamizadeh2022unetr,tang2022self}. For ultrasound or endoscopy videos the same idea applies across neighboring frames where temporal coherence is essential for reliable assessment \cite{ouyang2020video}. Practical measures such as 2.5D sampling, feature caching, and lightweight prompt updates can keep the adaptation efficient. The same principles can transfer to remote sensing and industrial inspection where robust grounding and cross view agreement are also important and where emerging video language models and generalist medical models provide compatible backbones \cite{lin2023video}.

\section{Conclusion}
This work introduced CoTBox-TTT, a model agnostic and plug and play test time training framework for medical visual question answering. The approach freezes backbone networks and adapts only small sets of soft prompts, grounds reasoning on explicit visual evidence with aligned bounding boxes, and aligns answers across the original image and a localized crop with an exponential moving average teacher. The framework requires no additional labels and leaves encoders and decoders unchanged, and it delivers consistent gains in open-ended recall and close-ended accuracy on VQA-RAD, SLAKE, and PathVQA across diverse backbones. These results indicate that lightweight test time adaptation can make medical vision language systems more reliable and interpretable without architectural changes or extra supervision.

{
    \small
    \bibliographystyle{ieeenat_fullname}
    \bibliography{main}
}


\end{document}